\documentclass[review]{elsarticle}
\usepackage{lineno,hyperref}
\usepackage{epstopdf}
\usepackage{bm}
\usepackage{makecell}
\usepackage{mathrsfs}
\usepackage{amsfonts}

\journal{Journal of \LaTeX\ Templates}

\begin{document}

\begin{frontmatter}

\title{Robust Facial Landmark Detection by Cross-order Cross-semantic Deep Network \tnoteref{mytitlenote}}
\tnotetext[mytitlenote]{Fully documented templates are available in the elsarticle package on \href{http://www.ctan.org/tex-archive/macros/latex/contrib/elsarticle}{CTAN}.}

\author[mymainaddress,mysecondaryaddress]{Jun Wan}
\author[mymainaddress,mythirdaddress]{Zhihui Lai}
\cortext[mycorrespondingauthor]{Corresponding author: Zhihui Lai}
\ead{lai\_zhi\_hui@163.com}
\author[mymainaddress,mythirdaddress]{Linlin Shen}
\author[mymainaddress,mythirdaddress]{Jie Zhou}
\cortext[mycorrespondingauthor]{Corresponding author: Jie Zhou}
\ead{ jie\_jpu@163.com}
\author[mymainaddress,mythirdaddress]{Can Gao}
\author[mysecondaryaddress]{Gang Xiao}
\author[mymainaddress]{Xianxu Hou}
\address[mymainaddress]{College of Computer Science and Software Engineering, Shenzhen University, Shenzhen, 518060, China}
\address[mysecondaryaddress]{School of Mathematics and Statistics, Hanshan Normal University, Chaozhou 521041, China}
\address[mythirdaddress]{Shenzhen Institute of Artificial Intelligence and Robotics for Society, Shenzhen, 518060, China.}
\begin{abstract}
	Recently, convolutional neural networks (CNNs)-based facial landmark detection methods have achieved great success. However, most of existing CNN-based facial landmark detection methods have not attempted to activate multiple correlated facial parts and learn different semantic features from them that they can not accurately model the relationships among the local details and can not fully explore more discriminative and fine semantic features, thus they suffer from partial occlusions and large pose variations. To address these problems, we propose a cross-order cross-semantic deep network (CCDN) to boost the semantic features learning for robust facial landmark detection. Specifically, a cross-order two-squeeze multi-excitation (CTM) module is proposed to introduce the cross-order channel correlations for more discriminative representations learning and multiple attention-specific part activation. Moreover, a novel cross-order cross-semantic (COCS) regularizer is designed to drive the network to learn cross-order cross-semantic features from different activation for facial landmark detection. It is interesting to show that by integrating the CTM module and COCS regularizer, the proposed CCDN can effectively activate and learn more fine and complementary cross-order cross-semantic features to improve the accuracy of facial landmark detection under extremely challenging scenarios. Experimental results on challenging benchmark datasets demonstrate the superiority of our CCDN over state-of-the-art facial landmark detection methods.
\end{abstract}

\end{frontmatter}

\begin{figure}[t]
	\begin{center}
		\includegraphics[width=0.8\linewidth]{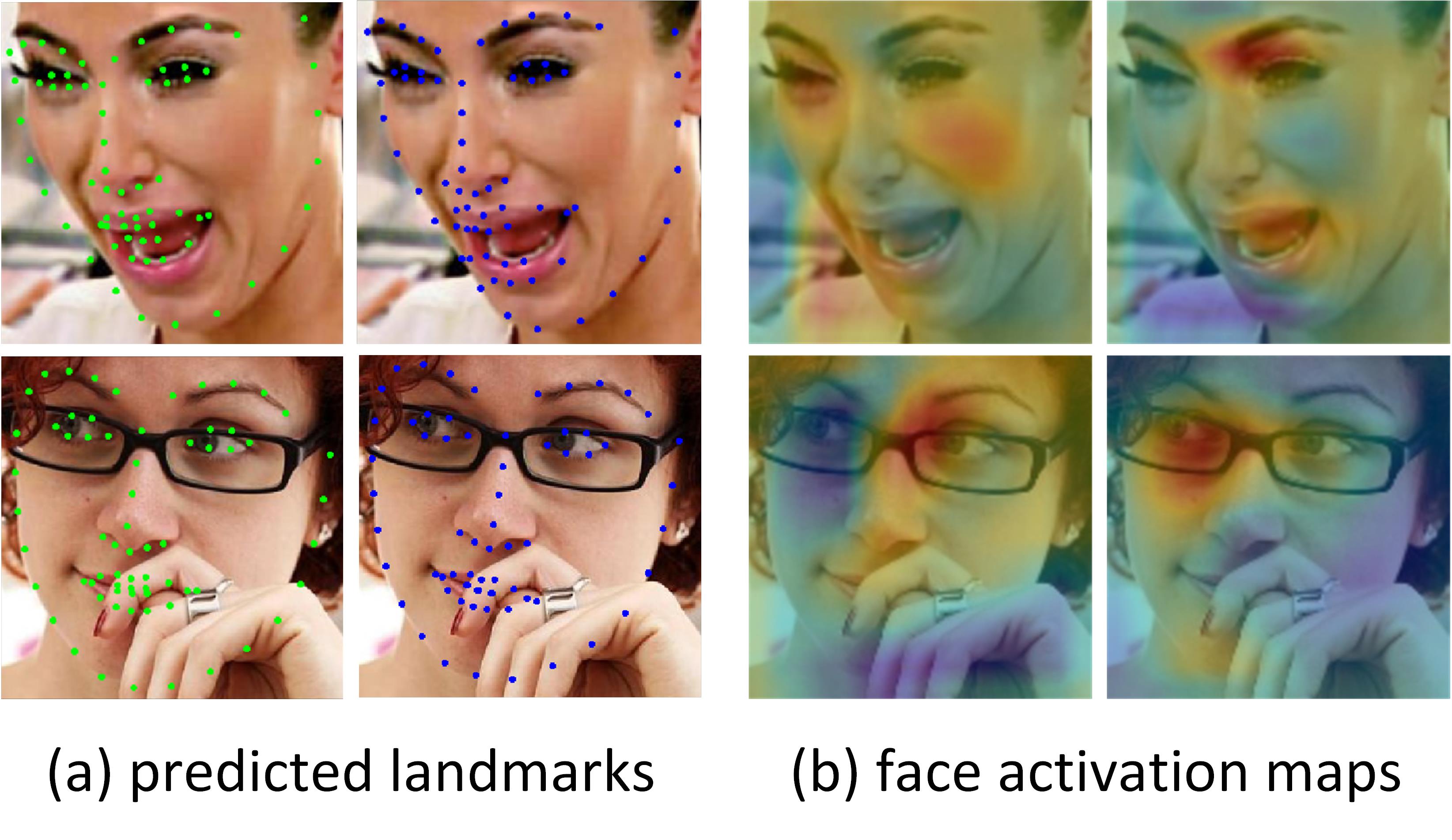}
	\end{center}
	\caption{(a) Our CCDN (blue dots) outperforms the state-of-the-art method (green dots). (b) From the face activation maps (two excitation blocks), we can see that CCDN can learn more fine and complementary semantic features, which can effectively capture the parts of interesting and suppress occluded ones, thus improving the accuracy of facial landmark detection with large pose variations and heavy occlusions.}
	\label{fig1}
\end{figure}
\section{INTRODUCTION}
Facial landmark detection, also known as face alignment, is a task to locate fiducial facial landmarks (eye corners, nose tip, etc.) in a face image, which can help achieve geometric image normalization and feature extraction. It becomes an indispensable part of facial analysis tasks such as face recognition \cite{Moghadam2018NonlinearAA}, face verification \cite{Xiong2020ECMLAE} and human-computer interaction \cite{Zheng2020DiscriminativeDM, Liu2020FacialER, Zhang2020AUD}. Recently, CNNs-based methods have been one of the mainstream approaches in facial landmark detection and achieve considerable performance on frontal faces. However, when suffering from large pose variations, heavy occlusions and complicated illuminations, CNN-based methods still cannot accurately detect landmarks.

The convolutional units in various layers of CNNs can actually pay more attention to parts of interest, i.e., behave as object detectors and landmark region detectors without any label information. Thus, CNNs-based facial landmark detection methods \cite{zhang2014facial, Wan2018FaceAB, Wu2018LookAB, Zhu2019RobustFL, Dong2018StyleAN, Liu2019SemanticAF,Kumar2020LUVLiFA, Chandran2020AD} are more robust to the variations in facial poses, expressions and occlusions. However, most CNNs-based facial landmark detection methods have not attempted to activate multiple correlated facial parts and learn different semantic features from them so that they cannot accurately model the differences between these correlated facial parts and the relationships among the local details in the correlated facial parts, i.e., they can not fully explore more discriminative and fine semantic features, thus the performance of the CNNs-based facial landmark detection method suffers from extremely large poses and heavy occlusions. For instance, the coordinate regression facial landmark detection methods \cite{zhang2014facial, Wu2018LookAB, Zhu2019RobustFL} learn features from the whole face images and then regress to the landmark coordinates, which drives the models to learn the whole facial features in a common/normal way that cannot accurately model the differences of local details and the relationships among local details. Also, the heatmap regression facial landmark detection methods \cite{Dong2018StyleAN, Liu2019SemanticAF,Kumar2020LUVLiFA, Chandran2020AD} generate a landmark heatmap for each landmark and then predict landmarks by traversing the corresponding landmark heatmaps. The region (for example, the mouth area) near the landmark largely determines the location of the predicted landmark, and the information of the other areas (eyes, eyebrow and forehead areas) has not yet been effectively encoded although deeper network structures are utilized to learn features with larger receptive fields and capture facial global constraints. Hence, as shown in Fig. \ref{fig1}, the above methods are not robust enough against large poses and partial occlusions. Furthermore, recent works have shown that the second-order statistics \cite{Gao2018GlobalSP, Dai2019SecondOrderAN, Wang2019DeepGG}, the part-specific semantic features \cite{luo2019cross, cai2016unified} and the feature selection mthods \cite{Li2015UnsupervisedFS, Li2020WeaklysupervisedSG, Gao2020Threeway} can help obtain more discriminative and robust features and are beneficial to many computer vision tasks. However, how to introduce the second-order statistics to activate parts of interest and then learn multiple more discriminative and fine attention-specific (part-based) semantic features for robust facial landmark detection are still open questions.
\begin{figure*}[!t]
	\centering
	\includegraphics[width=0.96\linewidth]{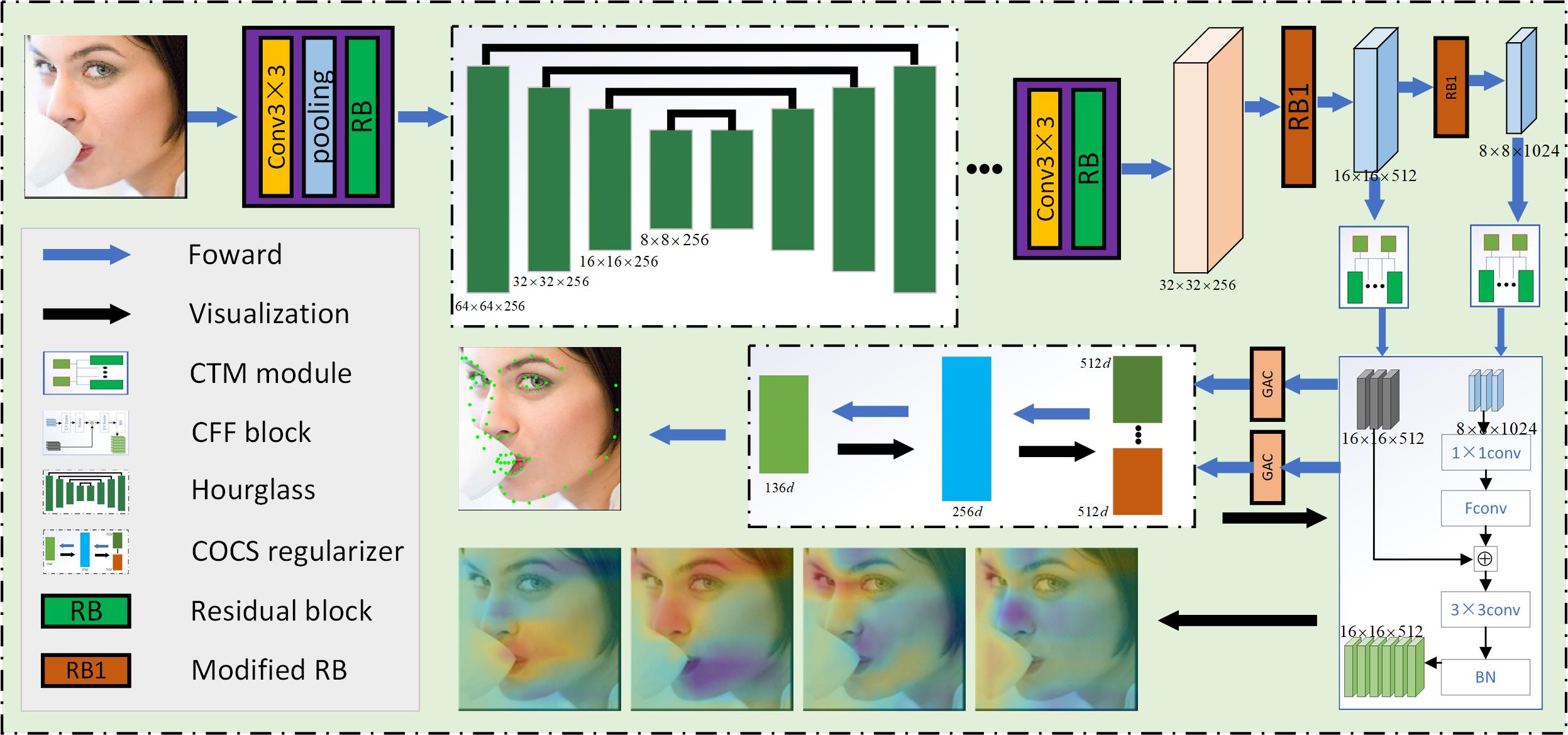}
	\centering
	\caption{ The overall architecture of our proposed cross-order cross-semantic deep network (CCDN). Firstly, four stacked hourglass networks are utilized to learn multi-scale features, which are then inputted into a simplified and modified SENet, i.e., the cross-order two-squeeze multi-excitation (CTM) module is proposed to generate multiple more discriminative attention-specific feature maps for activating more correlated facial parts. Finally, by integrating the CTM module and cross-order cross-semantic (COCS) regularizer into a novel CCDN with a seamless formulation, more discriminative and fine cross-order cross-semantic features can be learned for improving the accuracy of facial landmark detection. }
	\label{fig2}
\end{figure*}
To address the above problems, we propose a cross-order cross-semantic deep network (CCDN) to activate more correlated facial parts and learn more discriminative and fine cross-order cross-semantic features from them for more robust facial landmark detection. The overall architecture of the proposed CCDN is shown in Fig. \ref{fig2}. To be specific, we first propose a cross-order two-squeeze multi-excitation (CTM) module to generate multiple more discriminative attention-specific feature maps for activating more correlated facial parts. In the proposed CTM module, the cross-order channel correlations are introduced to selectively emphasize informative features and suppress less useful features by considering both the first-order and second-order statistics, thereby performing more effective feature re-calibration and generating more effective attention-specific feature maps. Then, a cross-order cross-semantic (COCS) regularizer is developed to guide the feature maps from different excitation blocks to represent different semantic meanings (i.e., activate different correlated facial parts) by maximizing the correlations between the features from the same excitation block, while de-correlating those from different excitation blocks. Finally, by integrating the CTM module and COCS regularizer via the proposed CCDN, more fine and complementary cross-order cross-semantic features can be learned for more robust facial landmark detection. Experimental results on benchmark datasets show that our approach obtains better robustness and higher accuracy than other state-of-the-art facial landmark detection methods.
\\\indent The main contributions of this work are summarized as follows:
\\\indent 1) With the well-designed CTM module, cross-order channel correlations can be introduced to perform more effective feature re-calibration and generate multiple more discriminative attention-specific feature maps, which helps learn more powerful cross-order cross-semantic features for robust facial landmark detection.
\\\indent 2) A COCS regularizer is designed to drive the network to learn the cross-order cross-semantic features from different excitation blocks. By exploring more fine and complementary semantic features, our method is able to enhance the robustness of facial landmark detection when facing large poses and heavy occlusions.
\\\indent 3) To the best of our knowledge, this is the first study to explore the cross-order cross-semantic features for handling facial landmark detection under challenging scenarios. By integrating the CTM module and COCS regularizer via a novel CCDN with a seamless formulation, our algorithm outperforms state-of-the-art methods on the benchmark datasets such as COFW \cite{Burgosartizzu2013Robust}, 300W \cite{Sagonas2016300FI}, AFLW \cite{Zhu2016UnconstrainedFA} and WFLW \cite{Wu2018LookAB}.
\\\indent The rest of the paper is organized as follows. Section \textbf{II} gives an overview of the related work. Section \textbf{III} shows the proposed method, including the CTM module and the COCS regularizer. A series of experiments are conducted to evaluate the performance of the proposed CCDN in Section \textbf{IV}. Finally, Section \textbf{V} concludes the paper. The symbols and their meanings are listed in Table \ref{tabsymbols}.
\begin{table}
	\caption{ Notations.}
	\vspace {1em}
	\centering
	\begin{tabular}{p{2cm}p{10cm}}
		\hline
		Notation &Meaning\\
		\hline
		$\mathbb{X}$ & the output feature map of a residual block ($\mathbb{X} \in {R^{J \times H \times d}}$).   \\
		$J$, $H$, $D$ & the width, height, and channels of $\mathbb{X}$.  \\
		${{{\bf{X}}_d}}$& the $d$-th channel of $\mathbb{X}$ (${{{\bf{X}}_d}} \in {R^{J \times H}}$).   \\
		$\bm{\kappa}^{st}$ &the first-order channel correlations ($\bm{\kappa}^{st} \in {R^{D}}$).
		\\
		$\bm{\kappa}^{nd}$ &the second-order channel correlations ($\bm{\kappa}^{nd} \in {R^{D}}$).\\
		$\hat{\mathbb{X}^{p}}$ & the output feature maps of excitation $p$. \\
		$\hat {\bf{Y}}$ & the final normalized covariance matrix (${\hat {\bf{Y}}} \in {R^{D \times D}}$). \\
		${\bf{f}}^p$ & the pooled features of the output of excitation $p$ (${\bf{f}}^p \in {R^D}$). \\
		$\bf{Q}$ & the correlation matrix of all pairwise excitation blocks. \\
		GAP & global average pooling. \\
		GCP & global covariance pooling. \\
		\hline
	\end{tabular}
	\label{tabsymbols}
\end{table}
\section{RELATED WORK}
During the past decades, rapid development has been made on facial landmark detection. Generally, most existing facial landmark detection methods can be divided into three groups: model-based methods, coordinate regression methods and heatmap regression methods.
\\\indent\textbf{Model-based methods.} Model-based methods learn parametric models (shape model \cite{cootes1995active}, appearance model \cite{cootes2001active} or part model \cite{Cristinacce2006FeatureDA}) from labeled datasets and leverage the principal component analysis (PCA) to model and constrain the shape variation to improve facial landmark detection. However, these methods are sensitive to large poses and partial occlusions.
\\\indent\textbf{Coordinate Regression methods.} The coordinate regression methods directly learn the mapping from facial appearance features to the landmark coordinate vectors by using various models \cite{zhang2014facial, Wu2018LookAB, cao2014face, Xiao2016Robust, Weng2016Learning, mo2019face}. In the supervised descent method \cite{xiong2013supervised} and local binary features \cite{ren2014face}, the linear regressor and random forest are separately utilized to perform cascaded regression for enhancing the robustness against the variations in facial expressions, head poses and illuminations. With a memory unit that shares information across all levels, a convolutional recurrent neural network \cite{trigeorgis2016mnemonic} is introduced to learn more effective and robust task-based features to enhance its robustness to large poses and partial occlusions. In cascaded regresssion and de-occlusion \cite{wan2020robust}, a cascaded deep generative regression model is proposed to handle face de-occlusion problems and face alignment problems simultaneously, which can effectively locate occlusions and recover more genuine faces that can be further used to improve the accuracy of facial landmark detection. By paying more attention to the contribution of samples with small and medium-size errors, the wing loss function \cite{Feng2017WingLF} is proposed to improve the deep neural network training capability and help to locate more accurate landmarks. In the look-at-boundary \cite{Wu2018LookAB}, firstly, the stacked hourglass networks are used to generate more accurate facial boundary heatmaps by introducing the message passing layers and adversarial learning concept. With these more accurate boundary heatmaps, its robustness against partial occlusions can be enhanced. In occlusion-adaptive deep networks \cite{Zhu2019RobustFL}, the Resnet is utilized to construct the geometry-aware module, the distillation module and the low-rank learning module for overcoming the occlusion problem in face alignment. With the effective shape constraints and the favorable regression ability, the robustness of the coordinate regression methods have been enhanced. However, they usually regress from the whole facial features to landmark coordinates, which can not accurately model the detailed differences in local details and the relationships among local details, thus they are not robust enough against facial landmark detection due to large poses and partial occlusions.
\\\indent\textbf{Heatmap Regression methods.} This category of methods \cite{Yang2017StackedHN, Dong2018StyleAN, Liu2019SemanticAF, Kumar2020LUVLiFA,Chandran2020AD} predicts the landmarks by generating and traversing the landmark heatmaps. To address the facial landmark detection problems causing by variations in the image styles, a style aggregated network  \cite{Dong2018StyleAN} is proposed to transform the original face image to style-aggregated images that are then used to train a more complicated model for robust facial landmark detection. Liu et al. \cite{Liu2019SemanticAF} propose a novel probabilistic model that can effectively predict the semantically consistent landmarks, the ground-truth landmark can be updated and the predict landmarks become more accurate. In LUVLi \cite{Kumar2020LUVLiFA}, a novel framework is proposed to jointly estimate facial landmark locations, uncertainty and visibility, thus providing accurate uncertainty predictions and improving the performance of facial landmark detection. Chandran et al. \cite{Chandran2020AD} propose a novel, fully convolutional regional architecture to address face alignment on very high-resolution images. By automatically defining and focusing on regions of interest instead of considering the image holistically, this architecture can achieve superior performance across all resolutions from 256 to 4K. Since this kind of method can better encode the part constraints and context information, and effectively drive the network to focus on parts of interest for facial landmark detection, they have achieved state-of-the-art accuracy. However, they are unable to take full advantage of distant and correlated facial local details and can not accurately model the relationships among correlated local details, that they still suffer from large poses and heavy occlusions.
\\\indent Until now, most CNNs-based facial landmark detection methods can not fully explore more discriminative and fine semantic features, as they can not accurately activate the most correlated facial parts and model the differences between them, suffering from performance degradation under large poses and heavy occlusions. Hence, we propose a cross-order cross-semantic deep network by integrating the CTM module and COCS regularizer for more robust facial landmark detection.

\section{Robust Facial Landmark Detection by Cross-order Cross-semantic Deep Network}

In this section, we firstly elaborate on the proposed CTM module and then present the COCS regularizer. Finally, we illustrate the proposed CCDN.
\begin{figure*}[t]
	\begin{center}
		\includegraphics[width=0.96\linewidth]{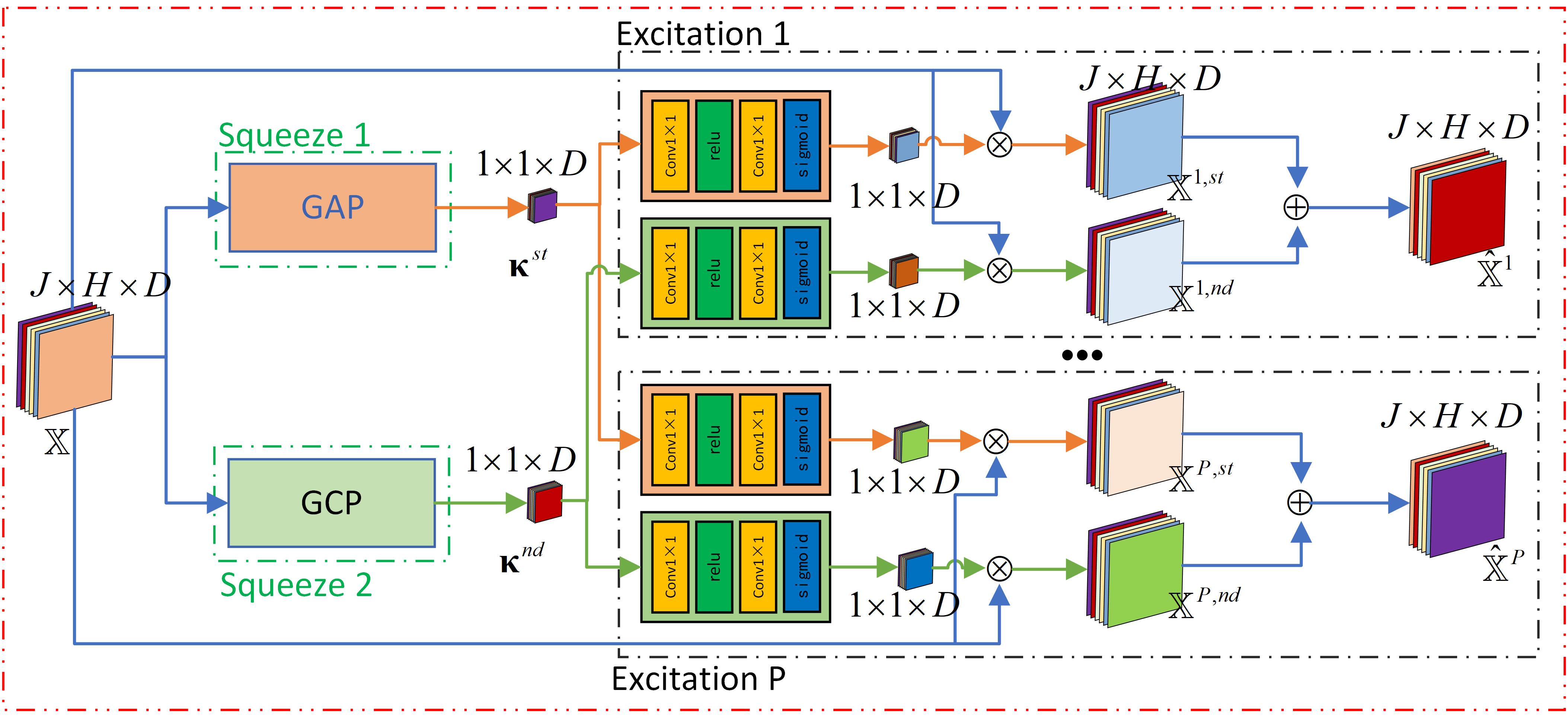}
	\end{center}
	\caption{The network structure of cross-order two-squeeze multi-excitation (CTM) module. By integrating the two-squeeze blocks and multi-excitation blocks into the proposed CTM module, the cross-order channel correlations containing both first-order and second-order statistics can be introduced to perform more effective channel-wise feature re-calibration and learn multiple attention-specific feature maps.}
	\label{figMTM}
\end{figure*}
\subsection{Cross-order Two-squeeze Multi-excitation Module}
Squeeze-and-Excitation Networks (SENet) \cite{hu2018squeeze} can effectively model channel-wise feature dependencies and perform dynamic channel-wise feature re-calibration. However, SENet ignores the statistics information higher than the first-order, thus hindering the discriminative ability of the network. Moreover, recent works \cite{Gao2018GlobalSP, Dai2019SecondOrderAN} have shown that second-order statistics are more discriminative and representative than first-order ones. Inspired by the above observations, we propose a cross-order two-squeeze multi-excitation module to perform more effective feature re-calibrating and multiple attention-specific feature maps generating by introducing the well-designed cross-order channel correlations. As shown in Fig. \ref{figMTM}, the cross-order channel correlations mainly contain the first-order and second-order channel correlations. With such proposed cross-order channel correlations, the two-squeeze blocks are able to learn more robust and discriminative representations for more effective feature re-calibrating and the multi-excitation blocks can help learn multiple attention-specific feature maps for each input image. Then, by integrating the two-squeeze blocks and the multi-excitation blocks via the proposed CTM module, our method is possible to activate parts of interest and ready to learn cross-order cross-semantic features.

\subsubsection{Two-squeeze blocks}
The two-squeeze blocks are proposed to introduce the cross-order channel correlations that contain both first-order channel correlations and second-order channel correlations. The \textbf{first-order channel correlations} can be introduced by global average pooling (GAP) operation. Specifically, let $\mathbb{X}$ denote the output feature map of a residual block, where $\mathbb{X} \in {R^{N \times D}}$ and $N = J \times H$. $J$, $H$ and $D$ are used to denote the width, height and channels of the feature maps, respectively. Let $\mathbb{X} = \left[ {{{{{\bf{X}}_1}}},{{{{\bf{X}}_2}}}, \cdots , {{{{\bf{X}}_d}}} ,\cdots,{{{{\bf{X}}_D}}}} \right]$, the $d$-th dimension of the first-order channel correlations ${{\bm{\kappa }}^{st}} \in {R^{D}}$ can be computed as:
\begin{equation}
	\bm{\kappa} _d^{st} = {{\rm{GAP}}}\left( {{{{{\bf{X}}_d}}}} \right) = \frac{1}{{J \times H}}\sum\limits_{j = 1}^J {\sum\limits_{h = 1}^H {{{{{\bf{X}}_d}}}\left( {j,h} \right)} }
\end{equation}

With the GAP operation in Eq. (1), the first-order channel correlations can be modeled and ready to be utilized for performing channel-wise feature re-calibration. The second-order statistics can be introduced by the global covariance pooling (GCP) \cite{Gao2018GlobalSP} operation, however, how to introduce the second-order statistics to the middle layers (i.e., the feature maps) of the network still pose a challenge. Inspired by recent works \cite{hu2018squeeze, luo2019cross, Dai2019SecondOrderAN}, we utilize the second-order statistics to model the channel-wise feature correlations and perform channel-wise feature re-calibration. Hence, the \textbf{second-order channel correlations} are introduced by the GCP operation that can be achieved by the Newton-Schulz iteration \cite{li2018towards}. We use $\hat {\bf{Y}}$ to denote the final normalized covariance matrix and  $\hat {\bf{Y}} = \left[ {{\bf{y}_1},{\bf{y}_2}, \cdots ,{\bf{y}_d}, \cdots ,{\bf{y}_D}} \right]$. ${\bf{y}_d}$ can be used to model the second-order channel-wise feature dependencies among the $d$-th channel and all channels. The $d$-th dimension of the second-order channel-wise feature dependencies $\bm{\kappa}^{nd}$ can be calculated as:
\begin{equation}
	\bm{\kappa} _d^{nd} = {\rm{GCP}}({{{\bf{X}}_d}}) = \frac{1}{D}\sum\nolimits_d {{\bf{y}_d}}
\end{equation}

With the GCP operation, the \textbf{second-order channel correlations} can be better utilized to model the channel-wise feature dependencies and help obtain more discriminative representations.
\subsubsection{Multi-excitation blocks}
With the two-squeeze blocks, the first-order and second-order channel correlations are modeled. Then they are utilized to perform feature re-calibration by gating mechanisms, in which a simple Sigmoid function can serve as a proper gating function. The whole process can be explored as :
\begin{equation}
	\begin{array}{l}
		\mathbb{X}^{p,ord} = {{\bf{s}}^{p,ord}}{\mathbb{X}}\\
		= {\rm{Sigmoid}} \left( {W_U^{p,ord} {\rm{ReLU}} \left({W_V^{p,ord} \bm{\kappa} ^{ord}} \right)} \right){\mathbb{X}}
	\end{array}
\end{equation}
where ${\rm{Sigmoid}}$ represents the sigmoid function, $ord \in \left\{ {st,nd} \right\}$ denotes the first-order or second-order statistics, ${\bf{s}}^{p,ord}$ denotes the scaling factor of excitation $p$ corresponding to the first-order or second-order channel correlations. ${W_U^{p,ord}}$ and ${W_V^{p,ord}}$ denote the parameters of $1 \times 1$ convolution layers, whose channel dimensions are set to ${D \mathord{\left/ {\vphantom {D 4}} \right. \kern-\nulldelimiterspace} 4}$ and $D$, respectively. $\mathbb{X}^{p,ord}$ represents re-calibrated feature maps with the first-order or second-order channel correlations corresponding to excitation $p$. Then, the final output of excitation $p$ can be illustrated as :
\begin{equation}
	\hat{{\mathbb{X}}^{p}} =\mathbb{X}^{p,st} + \mathbb{X}^{p,nd}
\end{equation}

With the proposed CTM module, the cross-order channel correlations can be used to perform more effective feature re-calibration and generate multiple more effective attention-specific feature maps for learning more discriminative and fine semantic features.

\subsection{Cross-order Cross-semantic regularizer}
The proposed CTM module can effectively perform feature re-calibration and generate multiple attention-specific feature maps by introducing the cross-order channel correlations. However, it is challenging to learn multiple more discriminative and fine semantic features from them, i.e., drive features from different excitation blocks to represent different semantic meanings. Metric learning \cite{sun2018multi} has been utilized to address the above problems by pulling positive features closer to the anchor and negative features away. However, it is hard to design an effective metric learning loss and how to optimize the loss is still a challenging problem. Hence, in this paper, we propose a cross-order cross-semantic regularizer to learn cross-order cross-semantic features by exploring the correlations between features from different layers and different excitation blocks. Specifically, we maximize feature correlations from the same excitation block and minimizing those from different excitation blocks.

Firstly, we perform GAP operation on the re-calibrated feature maps $\hat{\mathbb{X}_p}$ to obtain the corresponding pooled features $\bm{f}_p$ and $\bm{f}_p \in R^{D}$. Then, we normalize $\bm{f}_p$ by $l_2$-normalization. The pairwise correlations of two excitation blocks ${\bf{Q}}_{p,p'}$ can be formulated as:
\begin{equation}
	{{\bf{Q}}_{p,p'}} = \frac{1}{{{B^2}}}\sum {{\bf{F}}_p^T{{\bf{F}}_p}}
\end{equation}where $B$ is the batch size and ${{\bf{F}}_p} = \left[ {{{\bf{f}}_{p,1}},{{\bf{f}}_{p,2}}, \cdots ,{{\bf{f}}_{p,B}}} \right] \in {R^{D \times B}}$. ${\bf{Q}}$ denotes the correlations of all pairwise excitation blocks. By maximizing the correlations within the same excitation block (the main diagonal elements of ${\bf{Q}}$) and minimizing the correlations between different excitation blocks (the elements of ${\bf{Q}}$ except the main diagonal), the cross-semantic loss can be defined as follows:

\begin{equation}
	\mathop {\min }\limits_{\bf{Q}} \mathcal{L}\left( {\bf{Q}} \right) = \frac{{\left\| {\bf{Q}} \right\|_F^2 - \left\| {diag\left( {\bf{Q}} \right)} \right\|_2^2}}{{\left\| {diag\left( {\bf{Q}} \right)} \right\|_2^2}}
\end{equation}where $\| \cdot \|$ denotes the Frobenius norm and $diag({\bf{Q}})$ means the main diagonal elements of the matrix ${\bf{Q}}$. By optmizing the Eq. (6), more fine and complementary semantic features can be learned from the re-calibrated feature maps.

\begin{figure}[t]
	\begin{center}
		\includegraphics[width=0.92\linewidth]{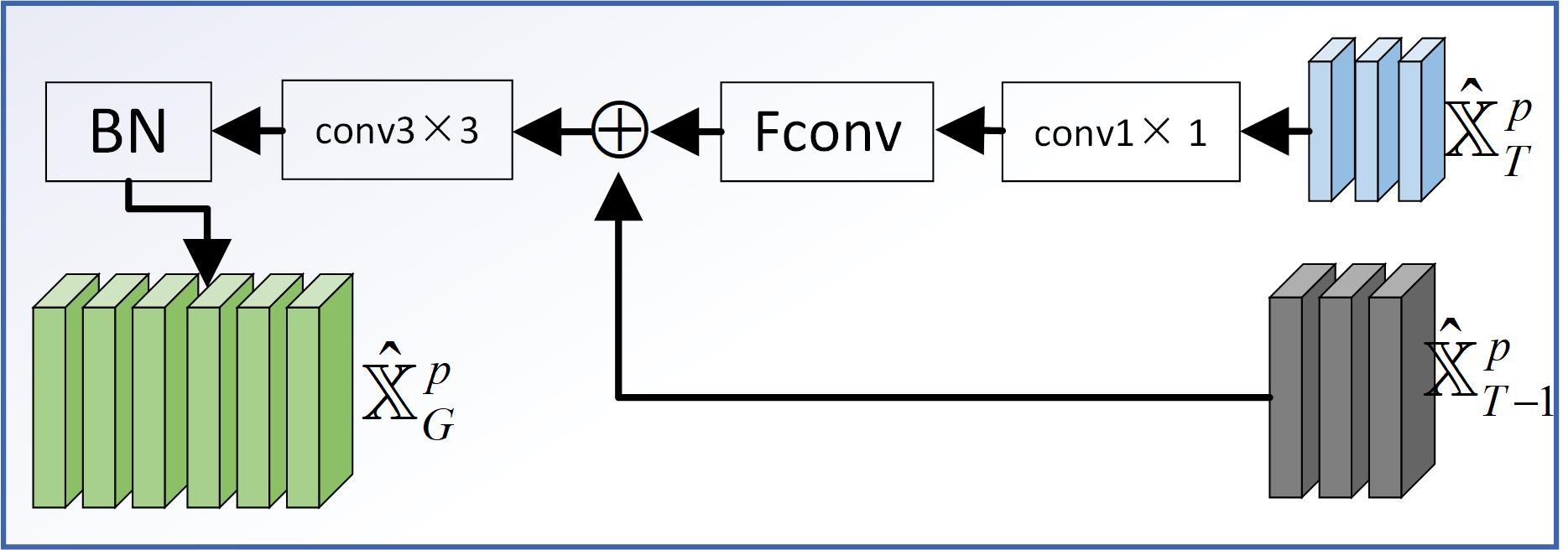}
	\end{center}
	\caption{With the well-designed cross-layer feature fusing (CFF) block, features from different layers can be aggregated to get more effective and discriminative representations, which are then optimized by the proposed cross-order cross-semantic regularizer to obtain cross-order cross-semantic features for robust facial landmark detection.}
	\label{figcff}
\end{figure}

As utilizing semantic features from different layers of CNNs has been shown to be beneficial to many vision tasks \cite{luo2019cross}, in this paper, we further fuse features from different layers by a cross-layer feature fusing (CFF) block as shown in Fig. \ref{figcff}. The CFF block utilizes the deconvolutional operation to upscale the size of the feature maps, which can reduce noise when compared with by other upscale operations such as bi-linear interpolation. The whole process of CFF block can be formulated as follows:

\begin{equation}
	\hat{\mathbb{X}_G^p}={\rm{BN}}({\bf{K}}_2* (\hat \mathbb{X}_{T - 1}^p+{\rm{Fconv}}({\bf{K}}_1*\hat{\mathbb{X}_T^p})))
\end{equation}where $*$, $\rm{BN}$ and $\rm{Fconv}$ are used to denote the convolutional, batch normalization (BN) and deconvolutional operations. $\hat{{\mathbb{X}}_T^p}$ and $\hat {\mathbb{X}}_{T - 1}^p$ denote the feature maps at layer $T$ and $T-1$, respectively. ${\bf{K}}_1$ denotes the $1 \times 1$ convolution kernel and ${\bf{K}}_2$ represents the $3 \times 3$ convolution kernel. With such a well-designed CFF block, features from different layers can be aggregated to learn more effective and discriminative representations. Hence, the final COCS regularizer loss can be formulated as follows:

\begin{equation}
	\mathop {\min }\limits_{\bf{Q}} E({\bf{Q}})= \mathcal{L}\left( {{{\bf{Q}}^T}} \right) + \mathcal{L}\left( {{{\bf{Q}}^{T - 1}}} \right) + \mathcal{L}\left( {{{\bf{Q}}^G}} \right)
\end{equation}where $\mathcal{L}\left( {{{\bf{Q}}^T}} \right)$, $\mathcal{L}\left( {{{\bf{Q}}^{T-1}}} \right)$ and  $\mathcal{L}\left( {{{\bf{Q}}^G}} \right)$ denote the cross-semantic loss at layers $T$, $T-1$ and of the fused feature maps, respectively. With such COCS regularizer loss, more fine and complementary cross-order cross-semantic features can be easily obtained, which helps improve the accuracy of facial landmark detection.
\subsection{Cross-order Cross-semantic Deep Network}
The proposed CTM module can perform effective feature re-calibration and generate multiple attention-specific feature maps by introducing the cross-order channel correlations. Then, the COCS regularizer is proposed to drive the network to learn different semantic features from different excitation blocks. Finally, the CTM module and COCS regularizer are integrated into a cross-order cross-semantic deep network with a seamless formulation to obtain more fine and complementary cross-order cross-semantic features that can be utilized to enhance the robustness of facial landmark detection under extremely large poses and heavy occlusions. The overall network structure of CCDN is shown in Fig. \ref{fig2} and the final objective function of our CCDN can be formulated as follows:

\begin{equation}
	\begin{array}{l}
		\mathop {\min }\limits_W \Phi \left( W \right) = \mathop {\min }\limits_W \sum\nolimits_i {} (\left\| {{{\bf{S}}_i^*} - CCDN\left( {W;IMG_i} \right)} \right\|_2^2\\
		{\kern 1pt} {\kern 1pt} {\kern 1pt} {\kern 1pt} {\kern 1pt} {\kern 1pt} {\kern 1pt} {\kern 1pt} {\kern 1pt} {\kern 1pt} {\kern 1pt} {\kern 1pt} {\kern 1pt} {\kern 1pt} {\kern 1pt} {\kern 1pt} {\kern 1pt} {\kern 1pt} {\kern 1pt} {\kern 1pt} {\kern 1pt} {\kern 1pt} {\kern 1pt} {\kern 1pt} {\kern 1pt} {\kern 1pt} {\kern 1pt} {\kern 1pt}  + {\gamma _1}\mathcal{L}\left( {{{\bf{Q}}^T}} \right) + {\gamma _2}\mathcal{L}\left( {{{\bf{Q}}^{T - 1}}} \right) + {\gamma _3}\mathcal{L}\left( {{{\bf{Q}}^G}} \right)
	\end{array}
\end{equation}where CCDN means the proposed approach. $W$ denotes the parameters of CCDN, ${\bf{S}}_i^*$ denotes the ground-truth shape of face image $IMG_i$, $\gamma_1$, $\gamma_2$ and $\gamma_3$ are hyper-parameters that balance the contribution of different costs. The first term is used to solve the mapping between the input facial features and the ground-truth landmarks. The other terms are used to obtain cross-order cross-semantic features and the regularization term is omitted for simplifying the formula. The optimization of CCDN is a typical end-to-end network training process under the supervision of Eq. (9). Hence, with the proposed CTM module and COCS regularizer, our CCDN is more robust against extremely large poses and heavy occlusions.
\section{Experiments}
In this section, we firstly introduce the evaluation settings including the datasets and the methods for comparison. Then, we compare our algorithm with state-of-the-art facial landmark detection methods on challenging benchmark datasets such as COFW \cite{Burgosartizzu2013Robust}, 300W \cite{Sagonas2016300FI}, AFLW \cite{Zhu2016UnconstrainedFA} and WFLW \cite{Wu2018LookAB}.
\subsection{Dataset and Implementation Details}

%
%

To validate the performance of the proposed method, we mainly introduce four challenging benchmark datasets with large variations in poses, expressions, occlusions and illuminations such as COFW \cite{Burgosartizzu2013Robust}, 300W \cite{Sagonas2016300FI}, AFLW \cite{Zhu2016UnconstrainedFA} and WFLW \cite{Wu2018LookAB}.
\\\indent\textbf{300W }(68 landmarks): It is a well-known competition dataset for facial landmark detection, and each face image is densely annotated with 68 landmarks. It consists of some present datasets including LFPW \cite{Zhu2012FaceDP}, AFW \cite{Belhumeur2011LocalizingPO}, Helen \cite{le2012interactive} and IBUG \cite{Sagonas2016300FI}. With a total of 3148 pictures, the training sets are made up of the training sets of AFW, LFPW and Helen while the testing set includes 689 images with such testing sets as IBUG, LFPW and Helen. The testing set is further divided  into three subsets: 1) Challenging subset (i.e., IBUG dataset). It contains 135 images that are collected from a more general ``in the wild" scenarios, and experiments on IBUG dataset are more challenging. 2) Common subset (554 images, including 224 images from LFPW test set and 330 images from Helen test set). 3) Fullset (689 images, containing the challenging subset and common subset).
\\\indent\textbf{COFW-68 }(68 landmarks): It is another very challenging dataset on occlusion issues which is published by Burgos-Artizzu et al. \cite{Burgosartizzu2013Robust}. It contains 1345 training images in which 845 images come from the LFPW \cite{Zhu2012FaceDP} dataset and the other images are heavily occluded. The testing set contains 507 face images with heavy occlusions, large pose variations and expression variations.
\\\indent\textbf{AFLW }(19 landmarks): It contains 25993 face images with jaw angles ranging from $ - {120^ \circ }$ to $ + {120^ \circ }$ and pitch angles ranging from $ - {90^ \circ }$ to $ + {90^ \circ }$. AFLW-full divides the 24386 images into two parts: 20000 for training and 4386 for testing. AFLW-frontal selects 1165 images out of the 4386 testing images to evaluate the alignment algorithm on frontal faces.

\indent\textbf{WFLW} (98 landmarks): It contains 10000 faces (7500 for training and 2500 for testing) with 98 landmarks. Apart from landmark annotation, WFLW also has rich attribute annotations (such as occlusion, pose, make-up, illumination, blur and expression.) that can be used for a comprehensive analysis of existing algorithms.

\textbf{Evaluation metric} Normalized Mean Error (NME) \cite{cao2014face} is commonly used to evaluate facial landmark detection methods. For 300W, NME normalized by the inter-pupil distance and the inter-ocular distance are utilized, separately. For COFW, NME normalized by the inter-pupil distance is used. For AFLW, we use NME normalized by the face size given by AFLW. For WFLW, NME normalized by inter-ocular distance is adopted.

\textbf{Implementation Details} In our experiments, we first use the four stacked hourglass networks as our backbone \cite{Yang2017StackedHN}. All the training and testing images are cropped and resized to $128 \times 128$ according to the provided bounding boxes. To perform data augmentation, we randomly sample the angle of rotation and the bounding box scale from Gaussian distribution. During the training process, we use the staircase function. The initial learning rate is $2.5 \times {10^{{\rm{ - 4}}}}$ which is decayed to $6.25 \times {10^{{\rm{ - 5}}}}$ after 100 epochs. The learning rate is divided by 2 and 2 at epoch 40 and 100, respectively. $p$ is set to 4. $\gamma_1$, $\gamma_2$ and $\gamma_3$ are set to 0.025, 0.01 and 0.05, resepectively. The CCDN is trained with Pytorch on 8 Nvidia Tesla V100 GPUs.
\begin{table}
	\caption{Comparisons with state-of-the-art methods on 300W dataset.}
	\begin{center}
		\begin{tabular}{c|ccc}
			\hline
			Method  &
			\begin{tabular}[c]{@{}c@{}}Common\\ Subset\end{tabular} & \begin{tabular}[c]{@{}c@{}}Challenging\\ Subset\end{tabular} & Fullset \\ \hline
			\multicolumn{4}{c}{Inter-pupil Normalization}                                                                                             \\ \hline
			3DDFA{$\rm _{CVPR16}$}\cite{Zhu2016FaceAA} &	5.53 &9.56 &6.31\\
			RAR{$\rm _{ECCV16}$}\cite{Xiao2016Robust}&	4.12	&8.35	&4.94\\
			Wing{$\rm _{CVPR18}$}\cite{Feng2017WingLF}&	3.27&	7.18&	4.04\\
			LAB{$\rm _{CVPR18}$}\cite{Wu2018LookAB}&	3.42	&6.98&	4.12\\
			BCNN-JDR{$\rm _{19'}$}\cite{Zhu2019BranchedCN} &	3.68	&7.16&	4.36\\
			Liu et al.{$\rm _{CVPR19}$}\cite{Liu2019SemanticAF} &	3.45&	6.38&	4.02\\
			ODN{$\rm _{CVPR19}$}\cite{Zhu2019RobustFL}	&3.56	&6.67&	4.17\\
			RCEN{$\rm _{19'}$}\cite{Lin2019region} &3.25 &6.70 &3.93 \\
			Chandran et al.{$\rm _{CVPR20}$}\cite{Chandran2020AD} &2.83 &7.04 &4.23 \\
			\hline
			Baseline &4.39 &7.41 &4.98 \\
			FCDN &3.64 &6.51 &4.20 \\
			\textbf{CCDN} &	\textbf{3.21}&	\textbf{6.02}&	\textbf{3.77}\\
			\hline
			\multicolumn{4}{c}{Inter-ocular Normalization}                                                       \\ \hline
			PCD-CNN{$\rm _{CVPR18}$}\cite{Kumar2018Disentangling3P} & 3.67 & 7.62      & 4.44    \\
			SAN{$\rm _{CVPR18}$}\cite{Dong2018StyleAN}           & 3.34    & 6.60      & 3.98    \\
			LAB{$\rm _{CVPR18}$}\cite{Wu2018LookAB}            & 2.98    & 5.19      & 3.49    \\
			DU-Net{$\rm _{ECCV18}$}\cite{Tang2018QuantizedDC}    & 2.90    & 5.15      & 3.35    \\
			HRNet{$\rm _{19'}$}\cite{Sun2019HighResolutionRF}               & 2.87    & 5.15      & 3.32    \\
			AWing{$\rm _{ICCV19}$}\cite{Wang2019AdaptiveWL}     & 2.72  & 4.52   & 3.07 \\
			LUVLi{$\rm _{CVPR20}$}\cite{Kumar2020LUVLiFA}       & {2.76}  & {5.16}    & {3.23}  \\ \hline
			\textbf{CCDN}                                        & 2.75 & 4.43 & 3.08 \\\hline
		\end{tabular}
	\end{center}
	\label{tab300w}
\end{table}

In the next section, we firstly compare our algorithm with the state-of-the-art methods, such as 3DDFA \cite{Zhu2016FaceAA}, RAR \cite{Xiao2016Robust}, Wing \cite{Feng2017WingLF}, LAB \cite{Wu2018LookAB}, DU-Net \cite{Tang2018QuantizedDC}, Liu et al. \cite{Liu2019SemanticAF}, ODN \cite{Zhu2019RobustFL}, HRNet \cite{Sun2019HighResolutionRF}, Chandran et al. \cite{Chandran2020AD} and LUVLi \cite{Kumar2020LUVLiFA}. When compared with those methods, we either use the released codes by the original authors or restore the original experiment, and both have achieved the expected effects in the corresponding papers. Moreover, we also compare our method with the \textbf{Baseline} (the Baseline uses four stacked hourglass networks to regress landmark coordinate vectors and its performance outperforms HGs \cite{Yang2017StackedHN}) and \textbf{FCDN} (the first-order statistics are only used to activate the correlated facial parts and learn fine and complementary cross-semantic features). Then, we show some exemplar results on facial landmark detection under large poses and partial occlusions and the corresponding results of performance evaluation in facial landmark detection, i.e., the NME and the cumulative error distribution (CED) curves.
\subsection{Comparison with State-of-the-Art methods}
\begin{figure*}[t]
	\begin{center}
		\includegraphics[width=0.8\linewidth]{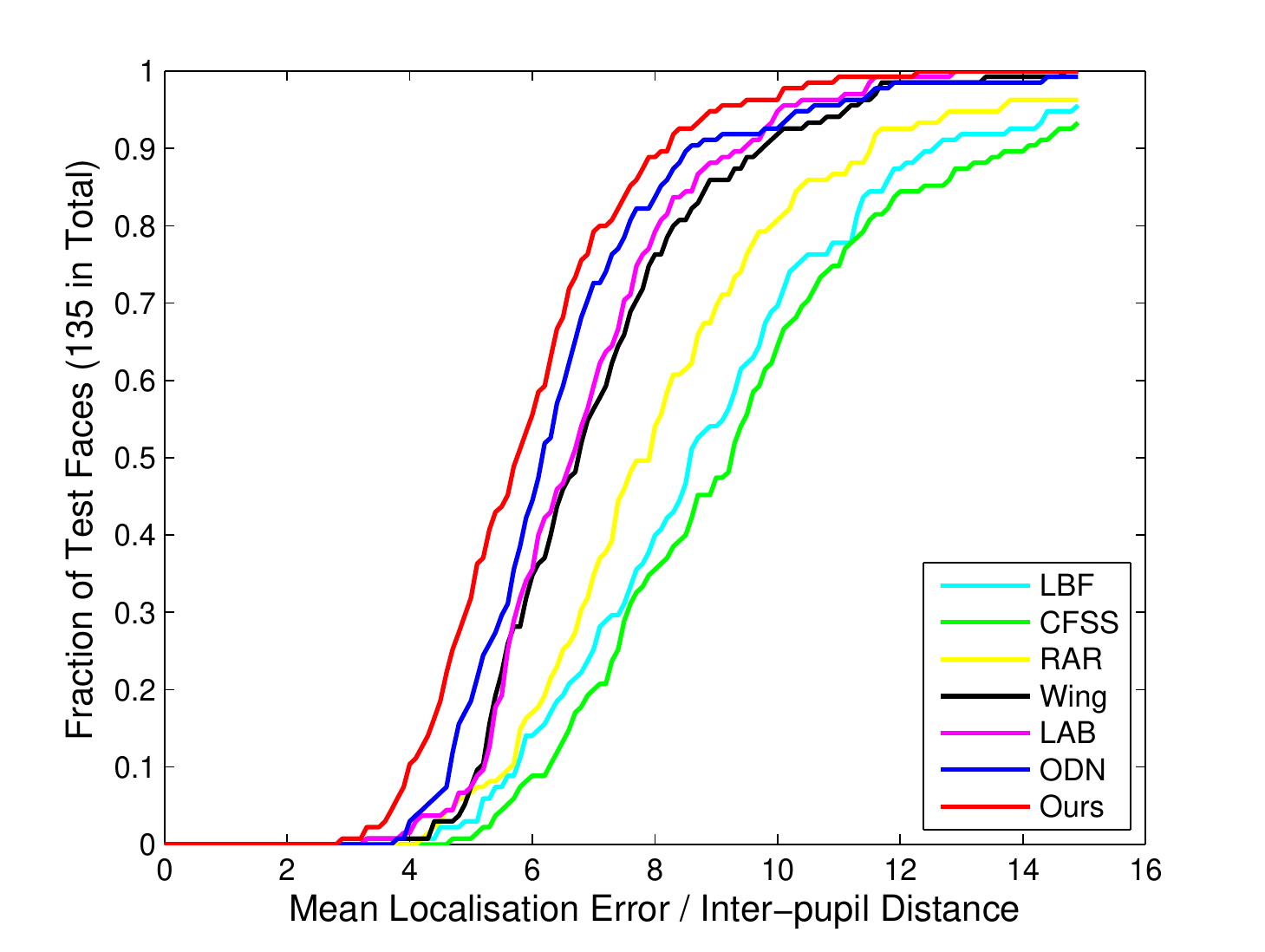}
	\end{center}
	\caption{Comparison of CED curves between our method and state-of-the-art methods on 300W Challenging Subset (68 landmarks). Our approach is more robust to partial occlusions and large poses than other methods. Best viewed in color.}
	\label{fig300w}
\end{figure*}
\begin{figure*}[t]
	\begin{center}
		\includegraphics[width=0.8\linewidth]{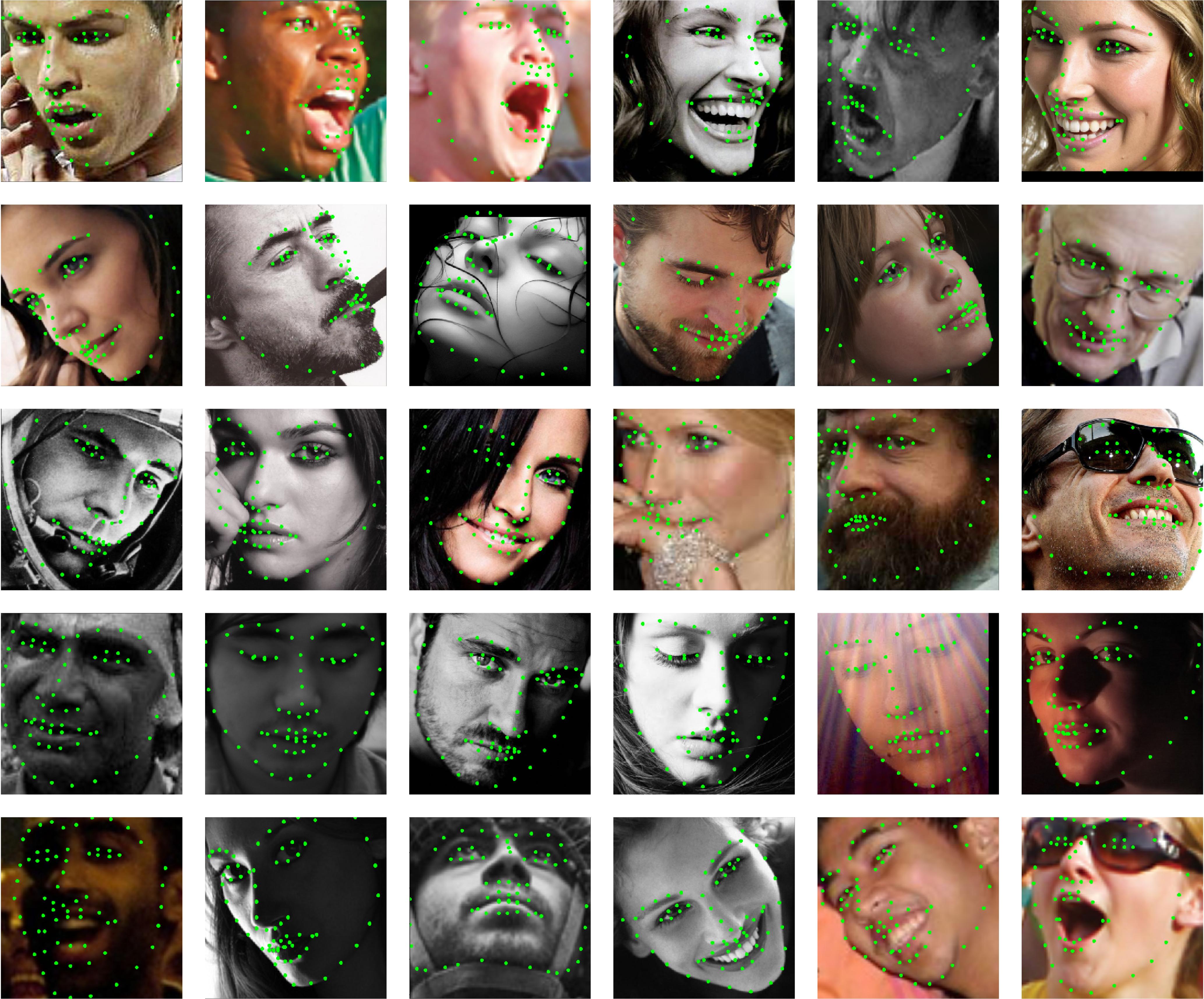}
	\end{center}
	\caption{Exemplar results on facial landmark detection under the variations in facial expressions (the 1st row), large poses (the 2nd row), partial occlusions (the 3rd row), illuminations (the 4th row) and complicated cases (the 5th row) on 300W dataset. With the proposed CCDN, our method is more robust to faces under challenging scenarios .}
	\label{figexample}
\end{figure*}
\subsubsection{Evaluation under Normal Circumstances}
Faces in the 300W common subset and full set have fewer variations in head poses, facial expressions and occlusions. Therefore, we evaluate the effectiveness of our method under normal circumstances on these two subsets. Table \ref{tab300w} displays the comparison of NME with state-of-the-art face alignment methods in detail. From Table \ref{tab300w}, we can see that our method achieves 3.21\% NME on the 300W common subset and 3.77\% NME on the 300W full set, which outperforms state-of-the-art methods on faces under normal circumstances. These results indicate that our CCDN can improve the accuracy of FLD under normal circumstances, which mainly because 1) the proposed CTM module can effectively perform feature re-calibration and generate multiple attention-specific feature maps by introducing the cross-order channel correlations; 2) the proposed COCS regularizer can drive the network to learn more fine and complementary cross-order cross-semantic features that possess more discriminative and representative probabilities, and 3) by integrating the CTM module and COCS regularizer into a novel CCDN with a seamless formulation, our method is able to the accuracy of facial landmark detection under normal circumstances.
\begin{table}
	\caption{Comparisons with state-of-the-art methods on COFW dataset. The error (NME) is normalized by the inter-pupil distance.}
	\begin{center}
		\begin{tabular}{p{5.6cm}p{0.8cm}p{1.0cm}}
			\hline
			Method & NME  & Failure \\
			\hline
			DRDA{$\rm _{CVPR16}$}\cite{Zhang2016OcclusionFreeFA}&	6.46&	6.00\\
			RAR{$\rm _{ECCV16}$}\cite{Xiao2016Robust}&	6.03&	4.14\\
			DAC-CSR{$\rm _{CVPR17}$}\cite{Feng2017DynamicAC}&	6.03&	4.73\\
			CAM{$\rm _{19'}$}\cite{Wan2019FaceAB}&	5.95&	3.94\\
			PCD-CNN{$\rm _{CVPR18}$}\cite{Kumar2018Disentangling3P} & 5.77 &3.73\\
			Wing{$\rm _{CVPR18}$}\cite{Feng2017WingLF}&	5.44&	3.75\\
			LAB{$\rm _{CVPR18}$}\cite{Wu2018LookAB}&	5.58&	2.76\\
			AWing{$\rm _{ICCV19}$}\cite{Wang2019AdaptiveWL}&	4.94&	0.99\\
			ODN{$\rm _{CVPR19}$}\cite{Zhu2019RobustFL}&	5.30&	-\\
			\hline
			Baseline &	6.17&	5.52\\
			FCDN &	5.32&	2.17\\
			\textbf{CCDN} &	\textbf{4.91}&	\textbf{1.12} \\
			\hline
		\end{tabular}
	\end{center}
	\label{tabcofw}
\end{table}
\subsubsection{Evaluation of Robustness against Occlusions}
Most state-of-the-art methods have got promising results on FLD under constrained environments. However, when face images suffer heavy occlusions and complicated illuminations, the accuracy of FLD will degrade greatly. In order to test the performance of our CCDN on faces with occlusions, we conduct the experiments on three heavy occluded benchmark datasets including the COFW dataset, 300W challenging subset and WFLW dataset.
\\\indent On the COFW dataset, the failure rate is defined by the percentage of test images with more than 10\% detection error. As illustrated in Table \ref{tabcofw}, we compare the proposed CCDN with other representative methods on the COFW dataset. Our CCDN boosts the NME to 4.91\% and the failure rate to 1.12\%, which outperforms the state-of-the-art methods.
\\\indent We also compare our approach against the state-of-the-art methods on the 300W challenging subset in Table \ref{tab300w} and Fig. \ref{fig300w}. As shown in Table \ref{tab300w}, our method achieves 6.02\% NME on the 300W challenging subset, which outperforms state-of-the-art methods on occluded faces. Furthermore, the cumulative error distribution(CED) curve in Fig. \ref{fig300w} and the exemplar results in facial landmark detection under complicated cases in Fig. \ref{figexample} also depict that our model achieves superior performance in comparison with other methods.
\\\indent The WFLW dataset contains complicated occluded subsets such as the Illumination subset, Make-Up Subset and Occlusion Subset. As shown in Table \ref{tabwflw}, our CCDN outperforms other state-of-the-art FLD methods.
\\\indent Hence, from the experimental results on the COFW dataset, 300W challenging subset and WFLW dataset, we can conclude that with the proposed CTM module and COCS regularizer, our CCDN can effectively activate multiple correlated facial parts and suppress the occluded parts, which help learn the cross-order cross-semantic features for enhancing its robustness against heavy occlusions.
\begin{table*}
	\caption{Comparisons with state-of-the-art methods on AFLW dataset. The error (NME) is normalized by face size.}
	\begin{center}
		\begin{tabular}{p{4.8cm}p{1.4cm}p{1.4cm}}
			\hline
			Method & full & frontal \\
			\hline
			CCL{$\rm _{CVPR16}$}\cite{Zhu2016UnconstrainedFA}&	2.72&	2.17\\
			TSR{$\rm _{CVPR17}$}\cite{Lv2017ADR}&	2.17&	-\\
			DAC-CSR{$\rm _{CVPR17}$}\cite{Feng2017DynamicAC}&	2.27&	1.81\\
			LLL{$\rm _{ICCV19}$}\cite{Robinson2019LaplaceLL}&	1.97&	-\\
			SAN{$\rm _{CVPR18}$}\cite{Dong2018StyleAN}&	1.91&	1.85\\
			DSRN{$\rm _{CVPR18}$}\cite{Miao2018DirectSR}&	1.86&	-\\
			LAB{$\rm _{CVPR18}$}\cite{Wu2018LookAB}&	1.85&	1.62\\
			HRNet{$\rm _{19'}$}\cite{Sun2019HighResolutionRF}&	1.85&	1.62\\
			ODN{$\rm _{CVPR19}$}\cite{Zhu2019RobustFL}&	1.63&	1.38\\
			Liu et al.{$\rm _{CVPR19}$}\cite{Liu2019SemanticAF} &1.60 &-\\
			LUVLi{$\rm _{CVPR20}$}\cite{Kumar2020LUVLiFA} &1.39 &1.19\\
			\hline
			Baseline &2.46 &1.90 \\
			FCDN &1.61 &1.36 \\
			\textbf{CCDN} &	\textbf{1.35}&	\textbf{1.14} \\
			\hline
		\end{tabular}
	\end{center}
	\vspace{-1em}
	\label{tabaflw}
\end{table*}
\begin{table*}
	\caption{Comparisons with state-of-the-art methods on WFLW dataset. The error (NME) is normalized by the inter-ocular distance.}
	\begin{center}
		\setlength{\leftskip}{-55pt}
		\begin{tabular}{p{3cm}p{1.3cm}p{1.3cm}p{1.5cm}p{1.7cm}p{1.5cm}p{1.3cm}p{1.2cm}}
			\hline
			Method  &
			\begin{tabular}[l]{@{}l@{}}Testset\\\end{tabular} & \begin{tabular}[l]{@{}l@{}}Pose\\ Subset\end{tabular} &
			\begin{tabular}[c]{@{}l@{}}Expression\\ Subset\end{tabular} &
			\begin{tabular}[c]{@{}l@{}}Illumination\\ Subset\end{tabular} &
			\begin{tabular}[c]{@{}l@{}}Make-Up\\ Subset\end{tabular} &
			\begin{tabular}[c]{@{}l@{}}Occlusion\\ Subset\end{tabular} &
			\begin{tabular}[c]{@{}l@{}}Blur\\ Subset\end{tabular}  \\ \hline
			CCFS{$\rm _{CVPR15}$}\cite{zhu2015face} &9.07 &21.36 &10.09 &8.30 &8.74 &11.76 &9.96 \\
			DVLN{$\rm _{CVPR17}$}\cite{wu2017leveraging} &6.08 &11.54 &6.78 &5.73 &5.98 &7.33 &6.88 \\
			LAB{$\rm _{CVPR18}$}\cite{Wu2018LookAB} &5.27 &10.24 &5.51 &5.23 &5.15 &6.79 &6.32 \\
			Wing{$\rm _{CVPR18}$}\cite{Feng2017WingLF} &5.11 &8.75 &5.36 &4.93 &5.41 &6.37 &5.81 \\
			HRNet{$\rm _{19'}$}\cite{Feng2017WingLF} &4.60 &7.86 &4.78 &4.57 &4.26 &5.42 & 5.36 \\
			AWing{$\rm _{ICCV19}$}\cite{Feng2017WingLF} &4.36 &7.38 &4.58 &4.32 &4.27 &5.19 &4.96 \\
			LUVLi{$\rm _{CVPR20}$}\cite{Feng2017WingLF} &4.37 &7.56 &4.77 &4.30 &4.33 &5.29 &4.94 \\
			\hline
			Baseline &5.78 &9.47 &6.39 &5.83 &5.91 &7.07 &7.21 \\
			FCDN &4.86 &7.81 &5.13 &4.77 &4.77 &5.78 &5.34 \\
			\textbf{CCDN} &\textbf{4.29} &\textbf{7.22} &\textbf{4.71} &\textbf{4.34} &\textbf{4.21} &\textbf{5.25} &\textbf{4.88}\\
			\hline
		\end{tabular}
	\end{center}
	\label{tabwflw}
\end{table*}

\begin{figure}[t]
	\begin{center}
		\includegraphics[width=0.8\linewidth]{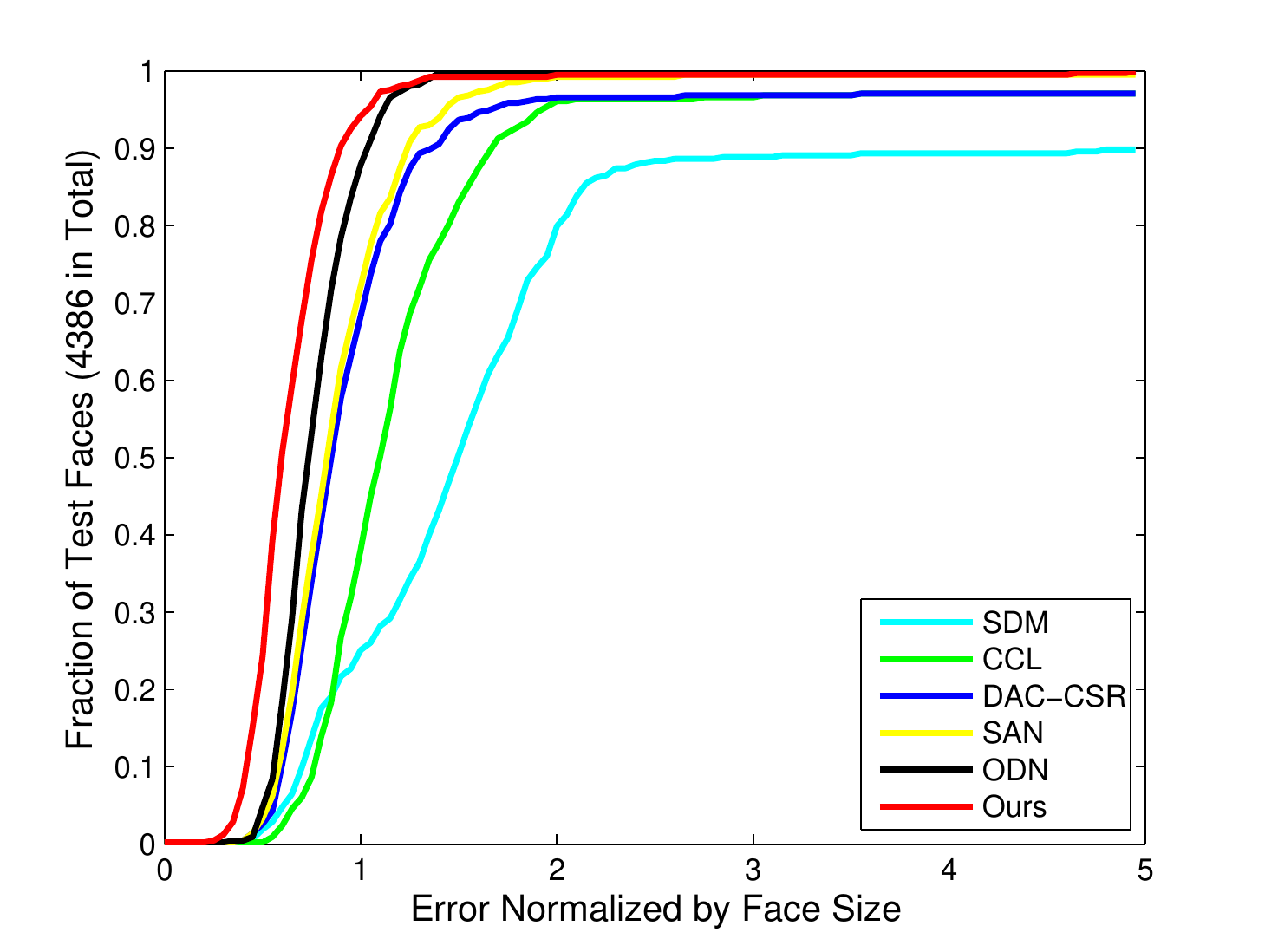}
	\end{center}
	\caption{Comparisons of CED curves of our method and state-of-the-art methods like SDM on AFLW-full dataset(19 landmarks). Our approach outperforms the other methods. Best viewed in color.}
	\label{figaflw}
\end{figure}
\subsubsection{Evaluation of Robustness against  Large Poses}
Face under large pose is another great challenge for FLD. To further evaluate the effectiveness of our proposed method under large poses, we carry out experiments on the AFLW dataset, 300W challenging subset and WFLW dataset. For the AFLW dataset, Table \ref{tabaflw} shows that our method achieves 1.35\% NME on the AFLW-full testing set and 1.14\% NME on the AFLW-frontal testing set, which outperforms the state-of-the-art methods. Furthermore, the CED curve in Fig. \ref{figaflw} also depicts that our model exceeds the other methods. For the WFLW dataset, the NME on the Pose Subset and Expression Subset surpasses the other methods. Hence, from the experimental results of the three datasets (see Table \ref{tab300w}, \ref{tabaflw}, \ref{tabwflw}, Fig. \ref{fig300w} and Fig. \ref{figaflw}), we can conclude that by incorporating the CTM module with the COCS regularizer, our CCDN can generate more effective multiple attention-specific feature maps, which helps to better activate parts of interest and learn more fine and complementary cross-order cross-semantic features, thus enhancing the robustness of our CCDN against the large pose and expression variations.
\begin{figure}[t]
	\begin{center}
		\includegraphics[width=0.8\linewidth]{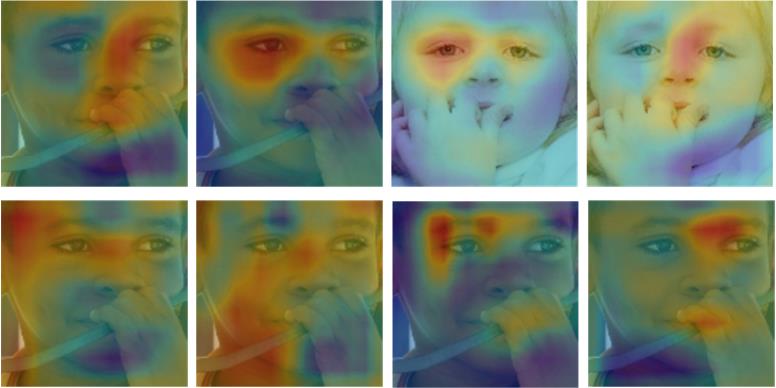}
	\end{center}
	\caption{ The first row represents the face activation maps with two excitation blocks corresponding to two face images. The second row represents the face activation maps with four excitation blocks. Compared with two excitation blocks, four excitation blocks can learn more detailed and robust face activation maps, that is, more effective semantic features.  Best viewed in color. }
	\label{figclcs}
\end{figure}

\begin{table}
	\caption{The effect (NME (\%)) of different $\gamma_1$, $\gamma_2$ and $\gamma_3$ values on the 300W challenging subset.}
	\vspace {1em}
	\centering
	\begin{tabular}{p{2cm}p{2cm}p{2cm}p{2cm}}
		\hline
		$\gamma_1$ &$\gamma_2$ & $\gamma_3$ &NME(\%)  \\
		\hline
		0.01 & 0.01 &0.01 &7.21 \\
		0.01 & 0.01 &0.025 &6.79 \\
		0.025 & 0.01 &0.025 &6.43 \\
		0.025 & 0.01 &0.05 &6.02 \\
		0.025 & 0.01 &0.1 &6.22 \\
		\hline
	\end{tabular}
	\label{tabhip}
\end{table}

\begin{table}
	\caption{Experimental results (NME (\%)) on the number of excitation blocks on benchmark datasets.}
	\vspace {1em}
	\centering
	\begin{tabular}{p{4.4cm}p{1cm}p{1cm}p{1cm}p{1cm}}
		\hline
		number &1 & 2 &3 &4  \\
		\hline
		300W-challenging & 7.41 &	6.35 &6.14 & 6.02 \\
		AFLW-full & 2.46 &	1.62 &1.48 & 1.35 \\
		WFLW-testset& 5.78 &	4.75 &4.44 & 4.29 \\
		\hline
	\end{tabular}
	\label{tabexc}
\end{table}

\subsection{Self Evaluations}
\textbf{Analysis of $\gamma_1$, $\gamma_2$ and $\gamma_3$.} We firstly calculate  $\mathcal{L}\left( {{{\bf{Q}}^T}} \right)$, $\mathcal{L}\left( {{{\bf{Q}}^{T-1}}} \right)$ and $\mathcal{L}\left( {{{\bf{Q}}^G}} \right)$ according to $\hat {\mathbb{X}_T^p}$, $\hat {\mathbb{X}_{T-1}^p}$ and $\hat {\mathbb{X}_{G}^p}$. $\mathcal{L}\left( {{{\bf{Q}}^T}} \right)$ and $\mathcal{L}\left( {{{{\bf{Q}}}^G}} \right)$ are closer to the output of networks than $\mathcal{L}\left( {{\bf{Q}^{T-1}}} \right)$, therefore, $\gamma_1$ and  $\gamma_3$ should be given large weights. Moreover, $\mathcal{L}\left( {{\bf{Q}^G}} \right)$ should be more important than $\mathcal{L}\left( {{\bf{Q}^T}} \right)$, hence, we have ${\gamma _2} \le {\gamma _1} \le {\gamma _3}$. We also conduct the corresponding experiments by using different $\gamma_1$, $\gamma_2$ and $\gamma_3$ values on the 300W challenging datasets. From the experimental results in Table \ref{tabhip}, we can find that $\gamma_1=0.025$, $\gamma_2=0.01$ and $\gamma_3=0.05$ are good choices to balance these four tasks.

\textbf{Number of excitation blocks.} The number of excitation blocks impacts the performance of facial landmark detection. Hence, we conduct experiments on numbers of excitation blocks on 300W dataset. Moreover, to visualize the cross-order cross-semantic features, we construct face activation maps by modifying the resized activation maps \cite{zhou2016learning}. Specifically, every two activation maps are constructed for each landmark (x and y coordinate). Then, by averaging activation maps of all landmarks, the face activation map is constructed. Fig. \ref{figclcs} displays the face activation maps with different excitation blocks. Also, Table \ref{tabexc} statistics the NME on 300W challenging subset, AFLW-full and WFLW-testset with different excitation blocks. From the experimental results from Fig. \ref{figclcs} and Table \ref{tabexc}, we can conclude that 1) face activation maps are more robust to partial occlusions and large poses as they can effectively activate the parts of interest and suppress the occluded ones; 2) the foreground (the red regions) of face activation maps is shown to cover large regions (including different facial parts such as eyebrow, eyes, nose and mouth area) so that they can effectively activate multiple correlated facial parts and help to better model the relationships between different facial local details; 3) face activation maps from the same layer complement each other -- they concentrate on different parts of faces which help to obtain more fine and complementary semantic features, and 4) with more excitation blocks, more fine cross-order cross-semantic features can be learned, which leads to more discriminative representations. Therefore, our CCDN is more robust to faces under the variations in poses, expressions, occlusions and illuminations.

\subsection{Ablation Study}
From the above experimental results, we can see that Our CCDN is more robust to faces under extremely large poses and heavy occlusions than state-of-the-art methods, which mainly because 1) we use several excitation blocks to learn more fine and complementary cross-order cross-semantic features that possess more representative and discriminative capabilities, and 2) by introducing the second-order statistics and COCS regularizer, the proposed CCDN can help activate more correlated facial parts and further improve the alignment accuracy. The former has been effectively verified in \textbf{Section 4.3}, we evaluate the latter in this ablation study by separately using the \textbf{Baseline} (the Baseline uses four stacked hourglass networks to regress landmark coordinate vectors and its performance outperforms HGs \cite{Yang2017StackedHN}), FCDN (the first-order statistics is only used to activate the correlated facial parts and learn fine and complementary cross-semantic features) and CCDN( the first-order and second-order statistics are both used). From experimental results in Tables \ref{tab300w}--\ref{tabwflw}, we can find that CCDN outperforms the Baseline and FCDN, and FCDN performs better than the Baseline, which indicates that 1) the second-order statistics can effectively help activate multiple correlated facial parts and learn more discriminative semantic features, and 2) by integrating the CTM module and COCS regularizer via the proposed CCDN, the cross-order cross-semantic features can be learned and utilized to enhance the robustness of facial landmark detection methods against heavy occlusions and large poses.

\subsection{Experimental results and discussions}
From the experimental results listed in Tables \ref{tab300w} -- \ref{tabexc} and the figures presented in previous subsections, we have the following observations and corresponding analyses.
\\\indent (1) CCDN and the heatmap regression methods (HGs \cite{Yang2017StackedHN}, SAN \cite{Dong2018StyleAN}, Liu et al. \cite{Liu2019SemanticAF} and LUVLi \cite{Kumar2020LUVLiFA}) enhance their robustness to the variations in facial poses, expressions and occlusions by effectively model the differences in facial local details and the correlations/relationships between different facial local details. The heatmap regression methods achieve this by utilizing the multi-scale features with larger receptive fields, while our method by proposing the cross-order cross-semantic deep network. However, our method outperforms the heatmap regression methods, which indicates that: 1) the well-designed cross-order two-squeeze multi-excitation module can effectively utilize the cross-order channel correlations to activate parts of interest and generate multiple attention-specific feature maps. 2) the proposed cross-order cross-semantic regularizer is able to drive the network to learn more fine and complementary cross-order cross-semantic features that are more robust to large poses and heavy occlusions from the generated multiple attention-specific feature maps. 3) by integrating the CTM module and COCS regularizer into a cross-order cross-semantic deep network with a seamless formulation, the robustness and accuracy of our method can be enhanced.
\\\indent (2) CCDN is similar to CCL \cite{Zhu2016UnconstrainedFA} and CAM \cite{Wan2019FaceAB} that firstly partition the optimization space of facial landmark detection into multiple domains of homogeneous descent and then fuse the regression results of relevant domains to further improve the accuracy of facial landmark detection. Compared to CCL and CAM, our method achieves this in an end-to-end way, i.e., by the proposed cross-order cross-semantic deep network. The final result of our method outperforms CCL and CAM, which indicates that our proposed CCDN can effectively activate parts of interest and learn more fine and complementary semantic features to enhance its robustness to the variations in large poses and heavy occlusions.
\\\indent (3) The second-order statistics can be utilized by the proposed CTM module to activate parts of interest and learn multiple attention-specific feature maps for learning the cross-order cross-semantic features. If we remove the second-order statistics, CCDN is equal to the FCDN. As shown in Table \ref{tab300w}--\ref{tabwflw}, CCDN exceeds the FCDN. This indicates that the second-order statistics can be used to effectively model the channel-wise feature dependencies and perform dynamic channel-wise feature re-calibration, which further helps activate parts of interest and learn multiple attention-specific feature maps for robust facial landmark detection.
\\\indent (4) The coordinate regression facial landmark detection methods \cite{zhang2014facial, Wu2018LookAB, Zhu2019RobustFL} learn features from the whole face images and then regress to the landmark coordinates, while our method firstly activates parts of interest and then learn multiple attention-specific semantic features to improve its accuracy. If we remove the learning of multiple attention-specific semantic features, CCDN is euqal to the Baseline. As shown in Table \ref{tab300w}--\ref{tabwflw}, our method outperforms the Baseline and the coordinate regression facial landmark detection methods \cite{zhang2014facial, Wu2018LookAB, Zhu2019RobustFL}. These indicate that by learning multiple attention-specific semantic features (more fine and complementary cross-order cross-semantic features), our method is able to achieve more accurate facial landmark detection.

(5) The more excitation blocks, the higher the accuracy of our method. However, more excitation blocks will consume more computational cost. From the experiment, we can find that utilizing four excitation blocks is a good choice that can effectively balance the accuracy and the computational cost.

\section{Conclusion}
Unconstrained facial landmark detection is still a very challenging topic due to large poses and partial occlusions. In this work, we present a cross-order cross-semantic deep network to address facial landmark detection under extremely large poses and heavy occlusions. By fusing the CTM module and the COCS regularizer with a seamless formulation, our CCDN is able to achieve more robust facial landmark detection. It is shown that the CTM module can effectively activate parts of interest and drive the network to learn multiple attention-specific feature maps, which can be further regularized by the COCS regularizer to learn different semantic features. With such more fine and complementary cross-order cross-semantic features, the robustness of our proposed CCDN to extremely large poses and heavy occlusions have been enhanced. It can also be found from the experiments that by integrating the second-order statistics into the activation of parts of interest and the learning of attention-specific features, more fine and complementary semantic features can be obtained, which further enhances the robustness of the proposed method in challenging scenarios.

\section{Acknowledgments}

This work is supported by the National Natural Science Foundation of China(Grant No.  62076164, 62002233, 61802267, 61976145 and 61806127 ), the Natural Science Foundation of Guangdong Province (Grant No. 2019A1515111121, 2018A030310451 and 2018A030310450), the Shenzhen Municipal Science and Technology Innovation Council (Grant No. JCYJ20180305124834854) and the China Postdoctoral Science Fundation (Grant No. 2020M672802).

\section*{References}

\bibliography{mybibfile}

\begin{thebibliography}{10}
\expandafter\ifx\csname url\endcsname\relax
  \def\url#1{\texttt{#1}}\fi
\expandafter\ifx\csname urlprefix\endcsname\relax\def\urlprefix{URL }\fi
\expandafter\ifx\csname href\endcsname\relax
  \def\href#1#2{#2} \def\path#1{#1}\fi

\bibitem{Moghadam2018NonlinearAA}
S.~M. Moghadam, S.~A. Seyyedsalehi, Nonlinear analysis and synthesis of video
  images using deep dynamic bottleneck neural networks for face recognition,
  Neural networks : the official journal of the International Neural Network
  Society 105 (2018) 304--315.

\bibitem{Xiong2020ECMLAE}
F.-R. Xiong, Y.~Xiao, Z.~Cao, Y.~Wang, J.~T. Zhou, J.~Wu, Ecml: An ensemble
  cascade metric-learning mechanism toward face verification., IEEE
  transactions on cybernetics PP.

\bibitem{Zheng2020DiscriminativeDM}
H.~Zheng, R.~Wang, W.~Ji, M.~Zong, W.~K. Wong, Z.~Lai, H.~Lv, Discriminative
  deep multi-task learning for facial expression recognition, Inf. Sci. 533
  (2020) 60--71.

\bibitem{Liu2020FacialER}
Y.~Liu, X.~Zhang, Y.~Lin, H.~Wang, Facial expression recognition via deep
  action units graph network based on psychological mechanism, IEEE
  Transactions on Cognitive and Developmental Systems 12 (2020) 311--322.

\bibitem{Zhang2020AUD}
F.~Zhang, T.~Zhang, Q.~Mao, C.~Xu, A unified deep model for joint facial
  expression recognition, face synthesis, and face alignment, IEEE Transactions
  on Image Processing 29 (2020) 6574--6589.

\bibitem{zhang2014facial}
Z.~Zhang, P.~Luo, C.~C. Loy, X.~Tang, Facial landmark detection by deep
  multi-task learning, in: European Conference on Computer Vision, Springer,
  2014, pp. 94--108.

\bibitem{Wan2018FaceAB}
J.~Wan, J.~Li, J.~Chang, Y.~Wu, Y.~Xiao, C.~Song, Face alignment by
  coarse-to-fine shape estimation, Chinese Journal of Electronics 27 (2018)
  1183--1191.

\bibitem{Wu2018LookAB}
W.~Wu, C.~Qian, S.~Yang, Q.~Wang, Y.~Cai, Q.~Zhou, Look at boundary: A
  boundary-aware face alignment algorithm, in: IEEE Conference on Computer
  Vision and Pattern Recognition, pp. 2129--2138.

\bibitem{Zhu2019RobustFL}
M.~Zhu, D.~Shi, M.~Zheng, M.~Sadiq, Robust facial landmark detection via
  occlusion-adaptive deep networks, in: IEEE Conference on Computer Vision and
  Pattern Recognition, pp. 3486--3496.

\bibitem{Dong2018StyleAN}
X.~Dong, Y.~Yan, W.~Ouyang, Y.~Yang, Style aggregated network for facial
  landmark detection, IEEE Conference on Computer Vision and Pattern
  Recognition (2018) 379--388.

\bibitem{Liu2019SemanticAF}
Z.~Liu, X.~Zhu, G.~Hu, H.~Guo, M.~Tang, Z.~Lei, N.~M. Robertson, J.~Wang,
  Semantic alignment: Finding semantically consistent ground-truth for facial
  landmark detection, in: IEEE Conference on Computer Vision and Pattern
  Recognition, pp. 3467--3476.

\bibitem{Kumar2020LUVLiFA}
A.~Kumar, T.~K. Marks, W.~Mou, Y.~Wang, M.~Jones, A.~Cherian, T.~Koike-Akino,
  X.~Liu, C.~Feng, Luvli face alignment: Estimating landmarks' location,
  uncertainty, and visibility likelihood, ArXiv abs/2004.02980.

\bibitem{Chandran2020AD}
P.~Chandran, D.~Bradley, M.~Gross, T.~Beeler, Attention-driven cropping for
  very high resolution facial landmark detection, 2020 IEEE/CVF Conference on
  Computer Vision and Pattern Recognition (2020) 5861--5870.

\bibitem{Gao2018GlobalSP}
Z.~Gao, J.~Xie, Q.~Wang, P.~Li, Global second-order pooling convolutional
  networks, in: IEEE Conference on Computer Vision and Pattern Recognition, pp.
  3024--3033.

\bibitem{Dai2019SecondOrderAN}
T.~Dai, J.~Cai, Y.~Zhang, S.-T. Xia, X.~P. Zhang, Second-order attention
  network for single image super-resolution, in: IEEE Conference on Computer
  Vision and Pattern Recognition, 2019.

\bibitem{Wang2019DeepGG}
Q.~Wang, P.~Li, Q.~Hu, P.~Zhu, W.~Zuo, Deep global generalized gaussian
  networks, in: IEEE Conference on Computer Vision and Pattern Recognition, pp.
  5080--5088.

\bibitem{luo2019cross}
W.~Luo, X.~Yang, X.~Mo, Y.~Lu, L.~S. Davis, J.~Li, J.~Yang, S.-N. Lim, Cross-x
  learning for fine-grained visual categorization, in: Proceedings of the IEEE
  International Conference on Computer Vision, 2019, pp. 8242--8251.

\bibitem{cai2016unified}
Z.~Cai, Q.~Fan, R.~S. Feris, N.~Vasconcelos, A unified multi-scale deep
  convolutional neural network for fast object detection, in: European
  conference on computer vision, Springer, 2016, pp. 354--370.

\bibitem{Li2015UnsupervisedFS}
Z.~Li, J.~Tang, Unsupervised feature selection via nonnegative spectral
  analysis and redundancy control, IEEE Transactions on Image Processing 24
  (2015) 5343--5355.

\bibitem{Li2020WeaklysupervisedSG}
Z.~Li, J.~Tang, L.~Zhang, J.~Yang, Weakly-supervised semantic guided hashing
  for social image retrieval, International Journal of Computer Vision (2020)
  1--14.

\bibitem{Gao2020Threeway}
C.~Gao, J.~Zhou, D.~Q. Miao, J.~J. Wen, X.~D. Yue, Three-way decision with
  co-training for partially labeled data, Information Sciences. (2020)
  doi:10.1016/j.ins.2020.08.104.

\bibitem{Burgosartizzu2013Robust}
X.~P. Burgosartizzu, P.~Perona, P.~Dollar, Robust face landmark estimation
  under occlusion, in: IEEE International Conference on Computer Vision, 2013,
  pp. 1513--1520.

\bibitem{Sagonas2016300FI}
C.~Sagonas, E.~Antonakos, G.~Tzimiropoulos, S.~P. Zafeiriou, M.~Pantic, 300
  faces in-the-wild challenge: database and results, Image Vision Comput. 47
  (2016) 3--18.

\bibitem{Zhu2016UnconstrainedFA}
S.~Zhu, C.~Li, C.~C. Loy, X.~Tang, Unconstrained face alignment via cascaded
  compositional learning, IEEE Conference on Computer Vision and Pattern
  Recognition (2016) 3409--3417.

\bibitem{cootes1995active}
T.~F. Cootes, C.~J. Taylor, D.~H. Cooper, J.~Graham, Active shape models-their
  training and application, Computer vision and image understanding 61~(1)
  (1995) 38--59.

\bibitem{cootes2001active}
T.~F. Cootes, G.~J. Edwards, C.~J. Taylor, Active appearance models, IEEE
  Transactions on pattern analysis and machine intelligence 23~(6) (2001)
  681--685.

\bibitem{Cristinacce2006FeatureDA}
D.~Cristinacce, T.~F. Cootes, Feature detection and tracking with constrained
  local models, in: British Machine Vision Conference, 2006.

\bibitem{cao2014face}
X.~Cao, Y.~Wei, F.~Wen, J.~Sun, Face alignment by explicit shape regression,
  International Journal of Computer Vision 107~(2) (2014) 177--190.

\bibitem{Xiao2016Robust}
S.~Xiao, J.~Feng, J.~Xing, H.~Lai, S.~Yan, A.~Kassim, Robust facial landmark
  detection via recurrent attentive-refinement networks, in: European
  Conference on Computer Vision, 2016, pp. 57--72.

\bibitem{Weng2016Learning}
R.~Weng, J.~Lu, Y.~P. Tan, J.~Zhou, Learning cascaded deep auto-encoder
  networks for face alignment, IEEE Transactions on Multimedia 18~(10) (2016)
  2066--2078.

\bibitem{mo2019face}
W.~Z. S.~Y. Huiyu~Mo, Leibo~Liu, S.~Wei, Face alignment with expression- and
  pose-based adaptive initialization, IEEE Transactions on Multimedia 21~(4)
  (2019) 943--956.

\bibitem{xiong2013supervised}
X.~Xiong, F.~De~la Torre, Supervised descent method and its applications to
  face alignment, in: IEEE Conference on Computer Vision and Pattern
  Recognition, 2013, pp. 532--539.

\bibitem{ren2014face}
S.~Ren, X.~Cao, Y.~Wei, J.~Sun, Face alignment at 3000 fps via regressing local
  binary features, in: IEEE Conference on Computer Vision and Pattern
  Recognition, 2014, pp. 1685--1692.

\bibitem{trigeorgis2016mnemonic}
G.~Trigeorgis, P.~Snape, M.~A. Nicolaou, E.~Antonakos, S.~Zafeiriou, Mnemonic
  descent method: A recurrent process applied for end-to-end face alignment,
  in: Proceedings of the IEEE Conference on Computer Vision and Pattern
  Recognition, 2016, pp. 4177--4187.

\bibitem{wan2020robust}
J.~Wan, J.~Li, Z.~Lai, B.~Du, L.~Zhang, Robust face alignment by cascaded
  regression and de-occlusion, Neural Networks 123 (2020) 261--272.

\bibitem{Feng2017WingLF}
Z.-H. Feng, J.~Kittler, M.~Awais, P.~Huber, X.~Wu, Wing loss for robust facial
  landmark localisation with convolutional neural networks, IEEE Conference on
  Computer Vision and Pattern Recognition (2017) 2235--2245.

\bibitem{Yang2017StackedHN}
J.~Yang, Q.~Liu, K.~Zhang, Stacked hourglass network for robust facial landmark
  localisation, IEEE Conference on Computer Vision and Pattern Recognition
  Workshops (2017) 2025--2033.

\bibitem{hu2018squeeze}
J.~Hu, L.~Shen, G.~Sun, Squeeze-and-excitation networks, in: Proceedings of the
  IEEE conference on computer vision and pattern recognition, 2018, pp.
  7132--7141.

\bibitem{li2018towards}
P.~Li, J.~Xie, Q.~Wang, Z.~Gao, Towards faster training of global covariance
  pooling networks by iterative matrix square root normalization, in:
  Proceedings of the IEEE Conference on Computer Vision and Pattern
  Recognition, 2018, pp. 947--955.

\bibitem{sun2018multi}
M.~Sun, Y.~Yuan, F.~Zhou, E.~Ding, Multi-attention multi-class constraint for
  fine-grained image recognition, in: Proceedings of the European Conference on
  Computer Vision (ECCV), 2018, pp. 805--821.

\bibitem{Zhu2012FaceDP}
X.~Zhu, D.~Ramanan, Face detection, pose estimation, and landmark localization
  in the wild, IEEE Conference on Computer Vision and Pattern Recognition
  (2012) 2879--2886.

\bibitem{Belhumeur2011LocalizingPO}
P.~N. Belhumeur, D.~W. Jacobs, D.~J. Kriegman, N.~Kumar, Localizing parts of
  faces using a consensus of exemplars, in: IEEE Conference on Computer Vision
  and Pattern Recognition, pp. 545--552.

\bibitem{le2012interactive}
V.~Le, J.~Brandt, Z.~Lin, L.~Bourdev, T.~Huang, Interactive facial feature
  localization, in: European Conference on Computer Vision, Springer, 2012, pp.
  679--692.

\bibitem{Zhu2016FaceAA}
X.~Zhu, Z.~Lei, X.~Liu, H.~Shi, S.~Z. Li, Face alignment across large poses: A
  3d solution, IEEE Conference on Computer Vision and Pattern Recognition
  (2016) 146--155.

\bibitem{Zhu2019BranchedCN}
M.~Zhu, D.~Shi, J.~Gao, Branched convolutional neural networks incorporated
  with jacobian deep regression for facial landmark detection, Neural networks
  : the official journal of the International Neural Network Society 118 (2019)
  127--139.

\bibitem{Lin2019region}
J.~W. C.~L. Xuxin~Lin, Yanyan~Liang, S.~Z. Li, Region-based context enhanced
  network for robust multiple face alignment, IEEE Transactions on Multimedia
  21~(12) (2019) 3053--3067.

\bibitem{Kumar2018Disentangling3P}
A.~Kumar, R.~Chellappa, Disentangling 3d pose in a dendritic cnn for
  unconstrained 2d face alignment, 2018 IEEE/CVF Conference on Computer Vision
  and Pattern Recognition (2018) 430--439.

\bibitem{Tang2018QuantizedDC}
Z.~Tang, X.~Peng, S.~Geng, L.~Wu, S.~Zhang, D.~N. Metaxas, Quantized densely
  connected u-nets for efficient landmark localization, in: Proceedings of the
  European Conference on Computer Vision (ECCV), 2018.

\bibitem{Sun2019HighResolutionRF}
K.~Sun, Y.~Zhao, B.~Jiang, T.~Cheng, B.~Xiao, D.~Liu, Y.~Mu, X.~Wang, W.~Liu,
  J.~Wang, High-resolution representations for labeling pixels and regions,
  ArXiv abs/1904.04514.

\bibitem{Wang2019AdaptiveWL}
X.~Wang, L.~Bo, F.~Li, Adaptive wing loss for robust face alignment via heatmap
  regression, 2019 IEEE/CVF International Conference on Computer Vision (ICCV)
  (2019) 6970--6980.

\bibitem{Zhang2016OcclusionFreeFA}
J.~Zhang, M.~Kan, S.~Shan, X.~Chen, Occlusion-free face alignment: Deep
  regression networks coupled with de-corrupt autoencoders, IEEE Conference on
  Computer Vision and Pattern Recognition (2016) 3428--3437.

\bibitem{Feng2017DynamicAC}
Z.-H. Feng, J.~Kittler, W.~J. Christmas, P.~Huber, X.~Wu, Dynamic
  attention-controlled cascaded shape regression exploiting training data
  augmentation and fuzzy-set sample weighting, IEEE Conference on Computer
  Vision and Pattern Recognition (2017) 3681--3690.

\bibitem{Wan2019FaceAB}
J.~Wan, J.~Li, J.~Chang, Y.~Wu, Y.~Xiao, X.~Li, H.~Zheng, Face alignment by
  component adaptive mechanism, Neurocomputing 329 (2019) 227--236.

\bibitem{Lv2017ADR}
J.-J. Lv, X.~Shao, J.~Xing, C.~Cheng, X.~Zhou, A deep regression architecture
  with two-stage re-initialization for high performance facial landmark
  detection, 2017 IEEE Conference on Computer Vision and Pattern Recognition
  (CVPR) (2017) 3691--3700.

\bibitem{Robinson2019LaplaceLL}
J.~P. Robinson, Y.~Li, N.~Zhang, Y.~Fu, S.~Tulyakov, Laplace landmark
  localization, 2019 IEEE/CVF International Conference on Computer Vision
  (ICCV) (2019) 10102--10111.

\bibitem{Miao2018DirectSR}
X.~Miao, X.~Zhen, X.~Liu, C.~Deng, V.~Athitsos, H.~Huang, Direct shape
  regression networks for end-to-end face alignment, 2018 IEEE/CVF Conference
  on Computer Vision and Pattern Recognition (2018) 5040--5049.

\bibitem{zhu2015face}
S.~Zhu, C.~Li, C.~Change~Loy, X.~Tang, Face alignment by coarse-to-fine shape
  searching, in: IEEE Conference on Computer Vision and Pattern Recognition,
  2015, pp. 4998--5006.

\bibitem{wu2017leveraging}
W.~Wu, S.~Yang, Leveraging intra and inter-dataset variations for robust face
  alignment, in: Proceedings of the IEEE conference on computer vision and
  pattern recognition workshops, 2017, pp. 150--159.

\bibitem{zhou2016learning}
B.~Zhou, A.~Khosla, A.~Lapedriza, A.~Oliva, A.~Torralba, Learning deep features
  for discriminative localization, in: Proceedings of the IEEE conference on
  computer vision and pattern recognition, 2016, pp. 2921--2929.

\end{thebibliography}

\end{document}